\documentclass[10pt,twocolumn,letterpaper]{article}

\usepackage[pagenumbers]{cvpr}              % To produce the CAMERA-READY version
\definecolor{cvprblue}{rgb}{0.21,0.49,0.74}
\usepackage[pagebackref,breaklinks,colorlinks,allcolors=cvprblue]{hyperref}
\usepackage{comment}
\usepackage{algorithm}
\usepackage{algpseudocode}
\usepackage{stmaryrd}
\usepackage[accsupp]{axessibility}
\def\paperID{CVSPORTS-65} % *** Enter the Paper ID here
\def\confName{CVPR}
\def\confYear{2025}

\title{Polar Coordinate-Based 2D Pose Prior with Neural Distance Field}

\author{Qi Gan$^1$
\and
Sao Mai Nguyen$^2$
\and
Eric Fenaux$^3$
\and
Stephan Clémençon$^1$
\and
Mounim El Yacoubi$^4$\\
\and
$^1$LTCI, Télécom Paris, Institut Polytechnique de Paris, France\\
$^2$U2IS, ENSTA Paris, Institut Polytechnique de Paris, France\\
$^3$Ef-e-science, France\\
$^4$SAMOVAR, Télécom SudParis, Institut Polytechnique de Paris, France\\
{\tt\small \{qi.gan, stephan.clemencon\}@telecom-paris.fr,}\\
{\tt\small \{nguyensmai, eric.fenaux\}@gmail.com, mounim.el\_yacoubi@telecom-sudparis.eu}
}

\begin{document}
\maketitle
\begin{abstract}
Human pose capture is essential for sports analysis, enabling precise evaluation of athletes' movements. While deep learning-based human pose estimation (HPE) models from RGB videos have achieved impressive performance on public datasets, their effectiveness in real-world sports scenarios is often hindered by motion blur, occlusions, and domain shifts across different pose representations. Fine-tuning these models can partially alleviate such challenges but typically requires large-scale annotated data and still struggles to generalize across diverse sports environments.
To address these limitations, we propose a 2D pose prior-guided refinement approach based on Neural Distance Fields (NDF). Unlike existing approaches that rely solely on angular representations of human poses, we introduce a polar coordinate-based representation that explicitly incorporates joint connection lengths, enabling a more accurate correction of erroneous pose estimations. Additionally, we define a novel non-geodesic distance metric that separates angular and radial discrepancies, which we demonstrate is better suited for polar representations than traditional geodesic distances. To mitigate data scarcity, we develop a gradient-based batch-projection augmentation strategy, which synthesizes realistic pose samples through iterative refinement. Our method is evaluated on a long jump dataset, demonstrating its ability to improve 2D pose estimation across multiple pose representations, making it robust across different domains. Experimental results show that our approach enhances pose plausibility while requiring only limited training data. Code is available at: \href{https://github.com/QGAN2019/polar-NDF}{https://github.com/QGAN2019/polar-NDF}.
\end{abstract}    
\section{Introduction}
\label{sec:introduction}

Human pose capture plays a vital role in sports analysis, particularly for the quantitative evaluation of athletic performance \cite{barris2008review}. By analyzing pose sequences, the technical intricacies of an athlete’s movements can be measured and assessed with high precision \cite{nekoui2020falcons, yeung2024autosoccerpose, gan2024human}. Despite its importance, accurately capturing human pose in sports scenarios remains a significant challenge.

\begin{figure*}[t]
  \centering
   \includegraphics[width=0.8\linewidth]{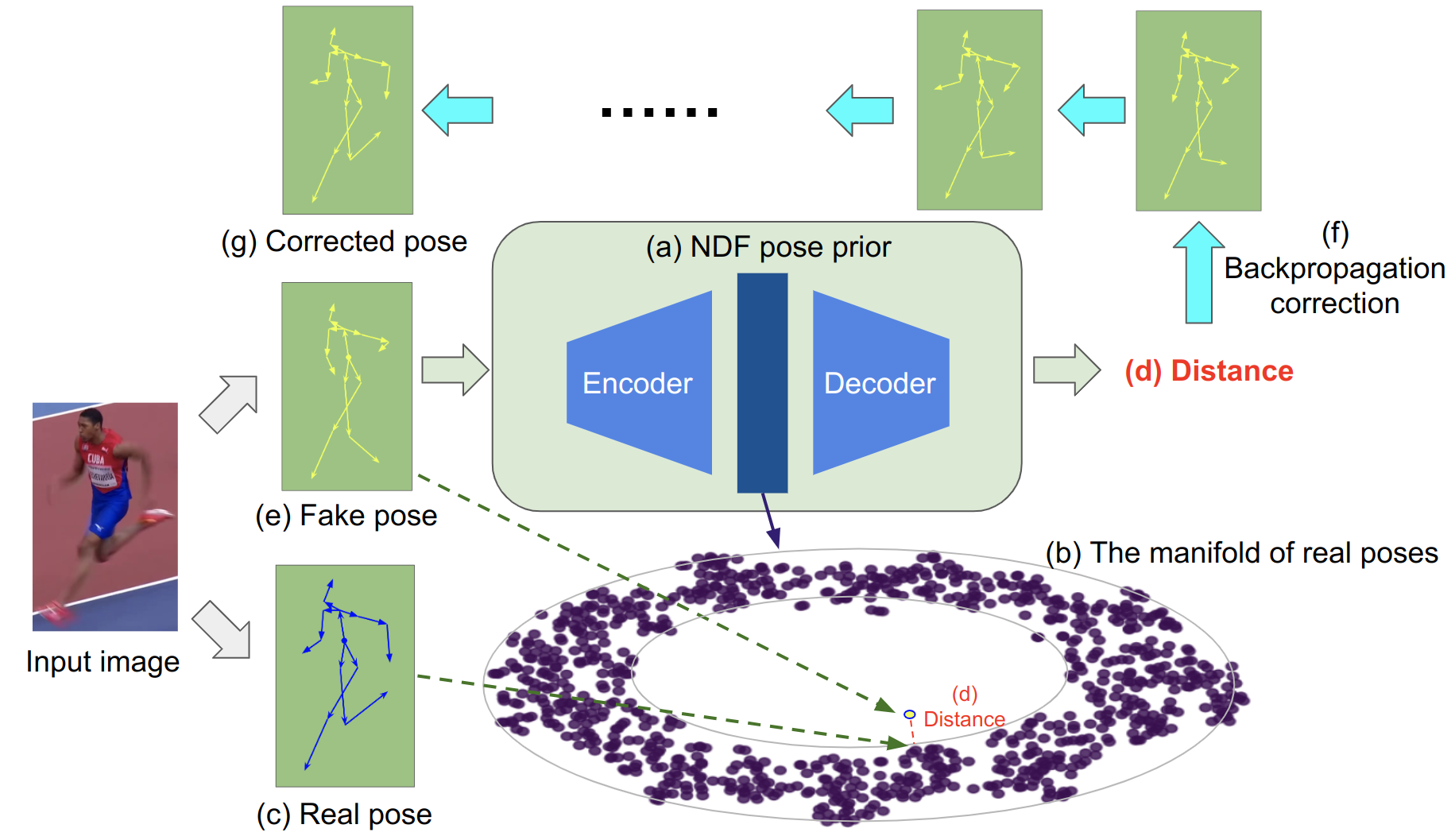}

   \caption{Overview of the method. We leverage a Neural Distance Field (NDF) network to learn the underlying pose prior manifold from real human poses. The (a) NDF network  learns (b) the zero-level-set manifold  (represented as deep blue points and visualized using t-SNE \cite{van2008visualizing}) of (c) real poses. The system estimates the (d) distance of (e) a fake pose in the embedded space (the yellow point) to the manifold. By performing several iterations of (f) backpropagation with the parameters of NDF fixed, the input fake pose is corrected, outputing (g) a corrected pose.}
   \label{fig:overview}
\end{figure*}

Traditionally, capturing human movement requires annotating human poses from videos frame-by-frame, which is time-consuming and costly, or dedicated motion capture systems requiring athletes to wear sensors, suits, or physical markers.  However, many sports applications demand fast, real-time feedback and non-intrusive pose acquisition. 
Recent advances in computer vision offer a promising alternative through deep neural network (DNN)-based human pose estimation \cite{Cao2019ITPAMI, fang2022alphapose, Sun2019PICCVPRC, xu2022vitpose}. Despite achieving human-level accuracy on public datasets \cite{lin2014microsoft}, these models often produce subpar results when applied to real-world sports videos. This performance gap stems from domain differences between the curated training datasets and the complex nature of sports footage. Sports videos commonly exhibit low frame rates, significant motion blur, dynamic lighting conditions, and frequent multi-person occlusions.
One strategy to address these issues is fine-tuning 2D Human Pose Estimation (HPE) models on domain-specific data. For instance, Ludwig \etal \cite{ludwig2021self} fine-tuned HRNet \cite{Sun2019PICCVPRC} using a self-supervised approach. However, this method is limited by the diversity of visual contexts in sports videos, including variations in background, lighting, uniforms, and camera angles—even within the same sport. Moreover, inconsistencies in pose representations across models further complicate result transferability \cite{Cao2019ITPAMI, fang2022alphapose, raychaudhuri2023prior}.
A more robust alternative involves post-correcting estimated poses using motion and structural constraints, such as temporal continuity \cite{zecha2018kinematic, zecha2019refining}, which has proven effective for correcting limb joint swaps. However, while temporal smoothing can fix local inconsistencies, it is insufficient for correcting broader detection errors.
A more effective approach to improving HPE results is to integrate prior knowledge of human pose into a dedicated pose prior model. While recent work on pose priors has predominantly focused on 3D human poses \cite{tiwari2022pose, aksan2019structured, xu2021h}, modeling 2D pose priors presents unique challenges. This is primarily because a single 3D pose can be projected into multiple valid 2D configurations, depending on the viewing angle or camera perspective.
POST \cite{raychaudhuri2023prior} successfully applied a 2D pose prior with Neural Distance Fields (NDF) \cite{chibane2020neural} for domain adaptation in 2D pose estimation using angular representations, where joint connection directions are encoded via their cosine and sine values. However, this angular representation ignores the lengths of joint connections (i.e., limb lengths), which are critical for evaluating the plausibility of a pose. A pose that is valid in terms of joint angles may become physically implausible if the limb lengths are altered. Therefore, angular 2D pose priors may face limitations when applied directly to certain HPE scenarios. Moreover, training an NDF-based pose prior typically requires large-scale datasets \cite{tiwari2022pose, raychaudhuri2023prior}.

To enable accurate and robust estimation of sports poses with limited training data, we propose a 2D pose prior-guided refinement method. Our approach extends prior work on neural distance field (NDF)-based pose priors \cite{tiwari2022pose, raychaudhuri2023prior}, which model the assumption that plausible human poses reside on a zero-level manifold, meaning all valid poses have zero distance to the manifold in the embedded space. In contrast, implausible poses lie outside this manifold, resulting in positive distances that reflect their deviation from physically realistic configurations.
To enable efficient correction of erroneous estimations, we introduce a novel pose representation and an accompanying distance metric. Unlike POST \cite{raychaudhuri2023prior}, our method employs a polar coordinate-based pose representation that explicitly incorporates joint connection lengths, enabling the pose prior to more effectively refine 2D estimations. In conjunction with this representation, we propose a non-geodesic distance metric that separately captures angular and radial discrepancies.
Leveraging this metric and representation, we further propose a gradient-based pose augmentation strategy that allows the NDF model to be trained with limited data. By backpropagating through arbitrary batches of poses, our method generates realistic synthetic samples that improve training efficiency. We validate our approach on a long jump dataset introduced by Gan \etal \cite{gan2024human}.
The contributions of this work include:
\begin{itemize}
    \item We propose a polar coordinate-based pose representation and a corresponding distance metric for NDF-based 2D human pose priors, along with a novel training scheme. This formulation enables the pose prior to refine both angular and radial errors in 2D pose estimations.
    \item We introduce a gradient-based data augmentation method that synthesizes new pose samples during training. This strategy allows the 2D NDF prior to be effectively learned from only a few video clips.
    \item We experimentally demonstrate that the learned 2D pose prior improves erroneous pose estimations across multiple 2D pose representations, enhancing robustness across different pose domains.
\end{itemize}

\section{Related works}
\label{sec:related_works}

\subsection{Human pose priors} 
Recent studies on human pose priors generally fall into two main categories. The first category involves Variational Autoencoder (VAE)-based approaches. For example, VPose \cite{pavlakos2019expressive} employs a variational prior to model human body poses, while HuMoR \cite{rempe2021humor} utilizes a conditional VAE to capture both temporal dynamics and body shape.
The second category leverages neural fields to represent human pose priors. Pose-NDF \cite{tiwari2022pose} models 3D human poses using a Neural Distance Field (NDF) \cite{chibane2020neural}, while H-NeRF \cite{xu2021h} learns 3D pose priors using Neural Radiance Fields (NeRF) \cite{mildenhall2021nerf}.
However, most existing methods focus on modeling 3D pose priors. In contrast, learning priors for 2D poses presents significantly greater challenges, as a single 3D human pose can project to many different 2D configurations depending on the camera viewpoint. Moreover, unlike 3D poses where joint connection lengths are fixed, 2D joint lengths vary continuously from zero to their full projected length, introducing additional variability.
To the best of our knowledge, the only existing work that models 2D human pose priors is POST \cite{raychaudhuri2023prior}, which employs a NDF. Their method represents 2D human poses solely through the orientations of joint connections, neglecting connection lengths. This omission is potentially problematic, as joint length is a critical factor in biomechanical plausibility, modifying connection lengths while preserving orientations can result in implausible poses. Furthermore, NDF-based priors typically require large-scale datasets for training \cite{raychaudhuri2023prior, tiwari2022pose}, limiting their applicability in data-scarce scenarios. 
Our approach adopts a polar coordinate-based representation to learn 2D pose priors using NDF, enabling the model to correct errors in both joint connection orientations and lengths. Additionally, we introduce a gradient projection-based data augmentation strategy that facilitates effective NDF learning even with limited training data.

\subsection{Sports pose estimation}
Applying HPE models directly to sports videos often yields suboptimal results due to the domain gap between everyday human poses and those specific to sports, as well as the degraded video quality caused by motion blur and occlusions. Two main approaches have been proposed to mitigate these issues.
The first is model fine-tuning. To address the challenge of limited annotated data, Ludwig \etal \cite{ludwig2021self} introduced two self-supervised learning schemes to enhance pose estimation performance, which were evaluated on long jump sequences.
The second approach involves correcting estimated poses. Fastovets \etal \cite{fastovets2013athlete} proposed an interactive, model-based generative framework for pose estimation in broadcast sports footage. Zecha \etal \cite{zecha2018kinematic, zecha2019refining} refined swimming pose estimates using temporal consistency. Dittakavi \etal \cite{dittakavi2022pose} introduced an explainable pose correction system based on an angle-likelihood mechanism, applied to yoga pose analysis.
In our work, we learn a 2D pose prior and address the data scarcity problem through a specialized gradient-based data augmentation strategy.
\section{Methodology}
\label{sec:methodology}
% The basic idea of NDF, why it is represented in polar, why bpj, how trained (briefly, more detailed should be provided in Introduction)
%Summary of this method
Our objective is to learn a 2D pose prior to correct erroneous pose estimations from sports videos (mostly broadcast competition videos) with a small amount of data. To achieve this goal, we propose a 2D polar coordinate pose NDF prior and a gradient-based data augmentation method.  In this section, we explain the details of the model and algorithm, as well as the training details.

\begin{figure}[t]
  \centering
   \includegraphics[width=0.7\linewidth]{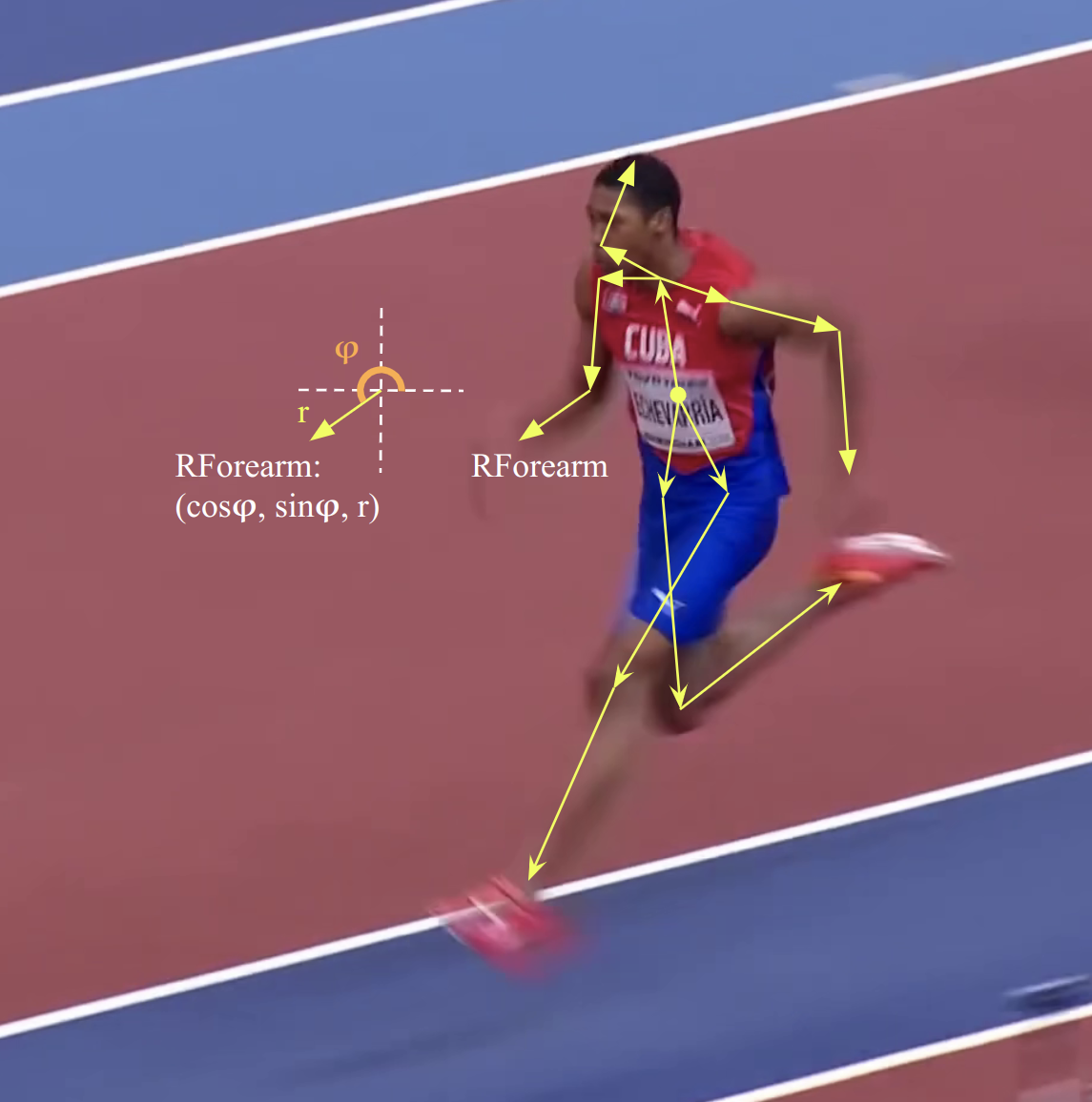}

   \caption{The 2D human pose representation used in this work is based on polar coordinates and consists of 15 joint connection vectors. Each vector is encoded by the cosine and sine of its orientation $\varphi$, along with its length (see the example of the 'RForearm' in the figure).}
   \label{fig:polar_representation}
\end{figure}

\subsection{Polar-coordinate representation NDF}
% The polar representation,  the network architecture, the loss term
In this work, we consider the skeleton representation of 2D human poses. As shown in Figure \ref{fig:polar_representation}, we represent a 2D pose skeleton with $J$ ($J=15$) vectors, while each vector $\boldsymbol{v}_i$ is represented by the cosine and sine values of its orientation angle ($\theta_i^{(1)}$ and $\theta_i^{(2)}$) and its length ($r_i$). Thus, the pose data could be represented as
$ \mathcal{X} = \{\boldsymbol{x} = (\boldsymbol{v}_1, ..., \boldsymbol{v}_J) |\boldsymbol{v}_i = (\theta_i^{(1)},\theta_i^{(2)},r_i)\in \mathbb{R}^3,||(\theta_i^{(1)},\theta_i^{(2)})||=1 \forall i \in[J]\}$. 

The neural distance field for 2D poses is the same as \cite{tiwari2022pose}, which is
$f^{ndf}:\mathcal{X} \to \mathcal{Y} \in\mathbb{R}_{\geq 0}$.
The neural distance field $f^{ndf}$ is modeled by a neural network in encoder-decoder structure $f^{ndf} := g^{dec}\circ g^{enc}$. The encoder $g^{enc}: \mathcal{X} \to \mathcal{Z} \in \mathbb{R}^{J\cdot L}$ is a hierarchical structured neural network \cite{aksan2019structured}, where $L$ is a hyperparameter and $J\cdot L$ is the dimension of the embedded feature space. The decoder $g^{dec}: \mathcal{Z} \to \mathcal{Y}$ is a simple MLP as in \cite{tiwari2022pose}.

Training of $f^{ndf}$ is regularized by L2 loss of distance predictions. For a training data $\mathcal{D}=\{(\boldsymbol{x},d)^{(i)}|i\in[N]\}$ which consists of real pose samples $(\boldsymbol{x}_{gt},d)^{(i)}$ and fake pose samples $(\boldsymbol{x}_{fk},d)^{(i)}$ (generated through the process described in \ref{subsec:fake_pose_generation}), the loss term for distance prediction is:
\begin{equation}
\begin{array}{cc}
     &  \mathcal{L}_{real} = \sum_{(\boldsymbol{x}_{gt},d)\in\mathcal{D}}||f^{ndf}(\boldsymbol{x}_{gt}) - d||_2,\\[10pt]
     & \mathcal{L}_{fake} = \sum_{(\boldsymbol{x}_{fk},d)\in\mathcal{D}}||f^{ndf}(\boldsymbol{x}_{fk}) - d||_2
\end{array}
\label{eq:loss_dist}
\end{equation}
In addition, we add a loss term to regularize the gradient of $f^{ndf}$ on real poses as 0 to ensure the NDF model converges better to real poses (represented as $\boldsymbol{x}_{gt}$):
\begin{equation}
    \mathcal{L}_{grad} = \sum_{(\boldsymbol{x}_{gt},d)\in\mathcal{D}}||\nabla f^{ndf}(\boldsymbol{x}_{gt})||_2.
    \label{eq:loss_grad}
\end{equation}
The total loss term is sum of Equation \ref{eq:loss_dist} and Equation \ref{eq:loss_grad}:
\begin{equation}
    \mathcal{L}_{total} = \mathcal{L}_{real} + \mathcal{L}_{fake} + \mathcal{L}_{grad}
\label{eq:loss_total}
\end{equation}

\subsection{Fake pose generation}
\label{subsec:fake_pose_generation}
To learn an NDF model, it is necessary to construct a dataset consisting of both real and fake poses, i.e. both samples on and outside the manifold. 
For real poses, it is straightforward to have zero distance because they are on the manifold. To augment the data, we flipped and interpolated the real poses. We then generate fake poses $\boldsymbol{x}_{fk}$ from erroneous estimates $\hat{\boldsymbol{x}}_{gt}$ from HPE models \cite{fang2022alphapose, Sun2019PICCVPRC, xu2022vitpose} and the corresponding ground truth pose $\boldsymbol{x}_{gt}$. First, we gather all detection errors into $\mathcal{E} = \{\boldsymbol{\varepsilon}^{(i)} = \hat{\boldsymbol{x}}_{gt}^{(i)} -\boldsymbol{x}_{gt}^{(i)} |i \in [N_{gt}] \}$, where $N_{gt}$ is the number of real poses in the training set. Then, fake poses are generated using the following equation:
\begin{equation}
\begin{array}{l}
    \boldsymbol{x}_{fk}^{(i)} = u\boldsymbol{\varepsilon}^{(j)} + \boldsymbol{x}_{gt}^{(k)}+\lambda T_{polar} (\boldsymbol{\xi}),\\[10pt]
    u \sim U(0,1), \boldsymbol{\xi} \sim \mathcal{N}(\boldsymbol{0},\boldsymbol{I}); i \in [M] ; j,k \in [N_{gt}]
\end{array}
\label{eq:fake_pose}
\end{equation}
where $M$ is the number of fake poses to be generated, $\boldsymbol{\xi}$ is random noise in Cartesian coordinates, $T_{polar}(\cdot)$ transforms poses in Cartesian coordinates to polar coordinates, and $\lambda$ is a small scaling factor. In Equation.\ref{eq:fake_pose}, we introduced a randomly sampled detection noise $\boldsymbol{\varepsilon}^{(j)}$ scaled by $u$. In this way, we ensure that the model focuses on learning the near-manifold regions that cover the real detection noises.

\subsection{Distance definition}
\label{subsec:dist_definition}
The computation of the distance of fake poses follows the procedure similar to the previous work \cite{tiwari2022pose, raychaudhuri2023prior}, except for the definition of distances. For a given fake pose $\boldsymbol{x}_{fk}$, we first find its $K$ nearest real poses $\{\boldsymbol{x}_{NN}^{(i)}|i\in [K] \}$ using a distance function $dist: (\mathcal{X},\mathcal{X})\to \mathcal{Y}\in \mathbb{R}_{\geq 0}$. A common definition of $dist$ is the geodesic distance weighted on each joint:
\begin{equation}
\begin{split}
    &dist(\boldsymbol{x}_{1}, \boldsymbol{x}_{2}) = \\
    &\sum_{j=1}^J w_j \sqrt{({r}_{j,1}{\theta}_{j,1}^{(1)} - r_{j,2}\theta_{j,2}^{(1)})^2 + ({r}_{j,1}{\theta}_{j,1}^{(2)} - r_{j,2}\theta_{j,2}^{(2)})^2},
\label{eq:geodesic_dist}
\end{split}
\end{equation}
where $w^{joint}_j$ is the pre-defined weight for joint $i$, $(\theta_{j,1}^{(1)}, \theta_{j,1}^{(2)}, r_{j,1})$ is the $j$th element of $\boldsymbol{x}_{1}$ and $(\theta_{j,2}^{(1)}, \theta_{j,2}^{(2)}, r_{j,2})$ is the $j$th element of $\boldsymbol{x}_{2}$. Another possible option for polar coordinate representations is to compute the angular and radial difference separately\cite{iwamoto2021polar, chen2022polar}. Following this idea, we propose an arc-radius-based distance:
\begin{equation}
\begin{split}
    &dist(\boldsymbol{x}_1, \boldsymbol{x}_2) = \\
    &\sum_{j=1}^J w_j (r_{j,1} \cdot\text{arccos}({\theta}_{j,1}^{(1)}\theta_{j,2}^{(1)} + {\theta}_{j,1}^{(2)}\theta_{j,2}^{(2)}) + |{r}_{j,1} - r_{j,2}|).
\label{eq:polar_dist}
\end{split}
\end{equation}
We experimentally show that Equation \ref{eq:polar_dist} is a better option than Equation \ref{eq:geodesic_dist} for polar coordinate representations (see Section \ref{subsec:exp_ablation} ). 
After the K-nearest real poses $\{\boldsymbol{x}_{NN}^{(i)}|i\in [K] \}$ are identified with $dist(\cdot)$, we compute the distance-weighted mean pose as the prior pose of $\tilde{\boldsymbol{x}}$ as:
\begin{equation}
\begin{array}{l}
    \boldsymbol{x}_{prior} = \sum_{i=1}^K w^{pose}_i \boldsymbol{x}_{NN}^{(i)}, \\[10pt]
    w^{pose}_i = \\
    (1 - dist(\boldsymbol{x}_{fk}, \boldsymbol{x}_{NN}^{(i)})/\sum_{k=1}^K dist(\boldsymbol{x}_{fk}, \boldsymbol{x}_{NN}^{(k)})) / (K-1),
\end{array}
\label{eq:weighted_dist}
\end{equation}
and re-calculate the distance $dist(\boldsymbol{x}_{fk}, \boldsymbol{x}_{prior})$ as the distance of $\boldsymbol{x}_{fk}$ to the manifold.

\subsection{Batch projection augmented training}
Training with only supervision from the distance of real/fake poses is not sufficiently effective when the training data are small, so we introduce batch projection-based data augmentation to enhance the training process. We generate a new sample ${\boldsymbol{x}}_{fk}^{bp}$ from a fake pose ${\boldsymbol{x}}_{fk}$ by gradient projection, which is similar to \cite{zhou2023learning}:
\begin{equation}
    {\boldsymbol{x}}_{fk}^{bp} = {\boldsymbol{x}}_{fk} - f^{ndf}( {\boldsymbol{x}}_{fk})\cdot \nabla f^{ndf}( {\boldsymbol{x}}_{fk}).
    \label{eq:bpj}
\end{equation}
This projection process is performed on the whole batch $B \in \mathcal{X}$ for specific iterations ($N^{bp}$) during each training epoch, where each batch consists of both real and fake samples, i.e. $B = \{B_{gt}, B_{fk}\}$. Once a sample is projected close to the manifold (determined with a threshold $\tau$), this sample will be excluded from the batch for the following projections. This iterative batch sample projection process supervises the model to better learn the fields from a fake pose sample to the manifold surface. The training process with batch projection data augmentation is summarized in Algorithm \ref{alg:bpj}.

\begin{algorithm}
\caption{Batch Projection Augmented Training}
\label{alg:bpj}
\begin{algorithmic}[1]
    \Require Training data $(X, Y)$, model $f^{ndf}(\cdot)$, learning rate $\eta$, projection iterations $N^{bp}$, threshold $\tau$
    \Ensure Updated model parameters $\theta$
    \For{each training epoch}
        \For{each batch $B = \{B_{gt}, B_{fk}\} \subset X$}
            \For{$n = 1$ to $N^{bp}$}
                \State Update $B_{fk} \gets B_{fk} - f^{ndf}(B_{fk}) \cdot \nabla f^{ndf}(B_{fk})$
                \For{each ${\boldsymbol{x}}_{fk} \in B_{fk}$}
                    \If{$ f^{ndf}({\boldsymbol{x}}_{fk}) < \tau$}
                        \State Remove ${\boldsymbol{x}}_{fk}$ from future projections
                        \State \textbf{break}
                    \EndIf
                \EndFor
            \EndFor
            \State Compute loss and update model $\theta$
        \EndFor
    \EndFor
    \State \textbf{return} Updated parameters $\theta$
\end{algorithmic}
\end{algorithm}

\subsection{Pose correction process}
We adapt the method in \cite{tiwari2022pose} to correct 2D poses with a trained NDF pose prior. Given an erroneous 2D pose estimate in Cartesian coordinate ${\boldsymbol{x}}_{fk}^{cart}$, by setting ${\boldsymbol{x}}_{fk}^{cart}$ learnable while the parameters of the NDF pose prior fixed, we use the following loss function to correct the pose:
\begin{equation}
    \mathcal{L}_{corr} = f^{ndf}(T_{polar}(\boldsymbol{x}_{fk}^{cart})).
\end{equation}
The correction iteration ends when the output distance of $f^{ndf}$ becomes smaller than a threshold, which is determined using the validation set.

\section{Experimental results}
\label{sec:experiments}
In this section, we first introduce the dataset and metric used for evaluating polar coordinate based 2D pose NDF prior (see Section \ref{subsec:exp_data_metric}). Then, we describe the implementation details of the experiments in Section \ref{subsec:exp_implementation}. Thereafter, we demonstrate the main results of pose correction with the trained prior (Section \ref{subsec:exp_main_results}), as well as a more detailed inspection into the joint-wise performance (Section \ref{subsec:exp_detailed_results}). Finally, we performed ablation studies on the proposed method in Section \ref{subsec:exp_ablation}.

\subsection{Dataset and metrics}
\label{subsec:exp_data_metric}
We evaluated the proposed pose prior using the long jump dataset introduced by Gan et al. \cite{gan2024human}. The dataset comprises 26 annotated clips of long jump performances by 22 athletes (11 men and 11 women), captured from broadcast footage of two World Championships. Each clip is recorded at a resolution of 1920×1080 and a frame rate of 25 frames per second. Ground-truth 2D pose annotations and foot–ground contact labels are provided. Using these contact flags, we segmented each clip to retain only the running and take-off phases where pose detection errors are most prevalent.
In addition to the ground-truth annotations, we generated pose estimates using three state-of-the-art HPE models: AlphaPose \cite{fang2022alphapose}, HRNet \cite{Sun2019PICCVPRC}, and ViTPose \cite{xu2022vitpose}, all applied with corrected bounding boxes detected by YOLOv6 \cite{liYOLOv6SingleStageObject2022} to improve input consistency.

Since the original dataset in \cite{gan2024human} does not include a predefined split, we manually defined a custom train–validation–test partition. Among the 26 long jump sequences, 6 sequences were randomly selected for the test set, 4 sequences for the validation set, and the remaining 16 sequences for training. From the training set, a subset of 8 sequences was selected to form the half-training set, and 4 sequences from that subset were further chosen as the quarter-training set. The total number of frames and pose annotations for each split is summarized in Table \ref{tab:data}.

We use Percentage of Correct Keypoints (PCK) as the evaluation metric in our experiments. The PCK metric involves a threshold $t$, which defines the reference distance as $t$ times the distance between the left shoulder and the right hip. A predicted joint is considered correct if its distance to the ground-truth joint is less than or equal to this reference distance. We evaluate performance at thresholds of $t = 0.1$, $0.2$, and $0.5$. According to \cite{ludwig2021self}, a threshold of $0.1$ approximately corresponds to a positional tolerance of $6$ cm.

\begin{table}
    \centering
    \begin{tabular}{@{}l c c@{}}
    \hline
    Data set&No. of sequences & No. of poses \\
    \hline
    & 16 (full) & 1249 \\
    %\cline{2-3}
    Training & 8 (half) & 638 \\
    %\cline{2-3}
    & 4 (quarter) & 280 \\
    \hline
    Validation & 4 & 315 \\
    \hline
    Test & 6 & 441 \\
    \hline
    Total & 26 & 2005\\
    \hline
    \end{tabular}
    \caption{Summary of data size and data splits.}
    \label{tab:data}
\end{table}

\subsection{Implementation details}
\label{subsec:exp_implementation}
The data were preprocessed to train the NDF pose prior. The experiments are primarily based on pose estimations from ViTPose \cite{xu2022vitpose}, HRNet \cite{Sun2019PICCVPRC}, and AlphaPose \cite{fang2022alphapose}. The pose skeleton definitions for ViTPose and HRNet follow the COCO WholeBody format \cite{jin2020whole}, while AlphaPose uses the Halpe26 format \cite{fang2022alphapose}. These skeletons were converted to a 17-keypoint format, consistent with the definition in \cite{ionescu2013human3}. The 2D poses were normalized such that all poses have unit height and their hip joints aligned at the origin.

For the training and validation sequences, we applied horizontal flipping and interpolation with a scaling factor of 5. We then generated 50 times more fake poses than the number of ground-truth poses, using the method described in Section \ref{subsec:fake_pose_generation}. These poses were subsequently transformed into either polar coordinates or angular coordinates (as in \cite{raychaudhuri2023prior}, for ablation studies). The distances between poses were computed using the algorithm detailed in Section \ref{subsec:dist_definition}. For each fake pose, we used FAISS \cite{johnson2019billion} to retrieve its 3 nearest-neighbor ground-truth poses for distance calculation.

We trained the NDF model using the Adam optimizer \cite{kingma2014adam} with a learning rate of 0.0001. We assigned equal weights to all loss components in Equation \ref{eq:loss_total}. Batch projection was conducted for 20 iterations per epoch across 20 total epochs. For comparison, when batch projection was not used, the model was trained for 400 epochs.

Pose correction on the test set was also performed using the Adam optimizer with a learning rate of 0.0001. During correction, each pose was refined over 100 iterations, or until the predicted distance from the NDF pose prior dropped below a threshold determined on the validation set.

\subsection{Overall evaluation}
\label{subsec:exp_main_results}
We conducted pose correction experiments using the trained NDF pose prior on the pose detections produced by AlphaPose \cite{fang2022alphapose}, HRNet \cite{Sun2019PICCVPRC}, and ViTPose \cite{xu2022vitpose}, respectively. The main results are presented in Table \ref{tab:main_result}. The findings indicate that the NDF pose prior substantially enhanced joint detection accuracy.

Evaluation of the raw pose detections reveals that the performances of the three models vary considerably. In general, ViTPose achieves the highest accuracy, while AlphaPose performs the worst under coarser thresholds (PCK@0.2 and PCK@0.5), and HRNet exhibits the lowest accuracy under stricter thresholds (PCK@0.05 and PCK@0.1).

The NDF pose prior consistently improved accuracy across all cases, demonstrating the method’s effectiveness and robustness. When comparing models trained on datasets of different sizes, although a larger training set yielded slightly better performance, the improvement margins remained modest. These results suggest that the proposed approach enables the model to learn effectively from small datasets without significant performance degradation. Despite the improved accuracy compared to raw detections, the overall gain is relatively limited. We attribute this primarily to convergence issues in the model, which are further discussed in Section \ref{subsec:further_studies}.

\begin{table*}
    \centering
    \begin{tabular}{@{}l c c c c c@{}}
    \hline
    Model & Method & PCK@0.05 & PCK@0.1 & PCK@0.2 & PCK@0.5 \\
    \hline
    AlphaPose & raw &44.88 &79.86 &91.32& 97.11 \\
    %\cline{2-5}
    & full &\textbf{45.11} &\underline{80.10} &\textbf{91.62} &\textbf{97.13} \\
    & half &\underline{45.08} &\textbf{80.22} &91.56 &\textbf{97.13} \\
    & quarter &44.94 &\underline{80.10} &\underline{91.57} &97.12 \\
    \hline
    HRNet & raw &41.18 &75.08 &94.12 &98.05  \\
    %\cline{2-5}
    & full &41.39 &\underline{76.35} &\underline{94.32} &\underline{98.12} \\
    & half &\underline{41.48} &76.20 &\textbf{94.37} &98.08 \\
    & quarter &\textbf{41.60} &\textbf{76.36} &94.30 &\textbf{98.13} \\
    \hline
    ViTPose & raw &51.18 &88.62 &98.07 &99.47 \\
    %\cline{2-5}
    & full &\underline{51.78} &\textbf{90.52} &\underline{98.19} &\underline{99.48} \\
    & half &51.35 &\textbf{90.52} &98.15 &\underline{99.48} \\
    & quarter &\textbf{51.95} &90.21 &\textbf{98.20} &\textbf{99.49} \\
    \hline
    \end{tabular}
    \caption{Evaluation of the pose correction performance using the trained pose prior. In the 'Method' column, \textit{raw} refers to the uncorrected pose estimations obtained directly from the models listed in the first column. \textit{Full}, \textit{half}, and \textit{quarter} indicate the respective proportions of the training set (as defined in Table \ref{tab:data}) used to train the pose prior model.
    }
    \label{tab:main_result}
\end{table*}

\subsection{Joint-wise correction}
\label{subsec:exp_detailed_results}
\begin{figure}[htbp]
    \centering
    \subfloat[PCK@0.05]{\includegraphics[width=0.8\linewidth]{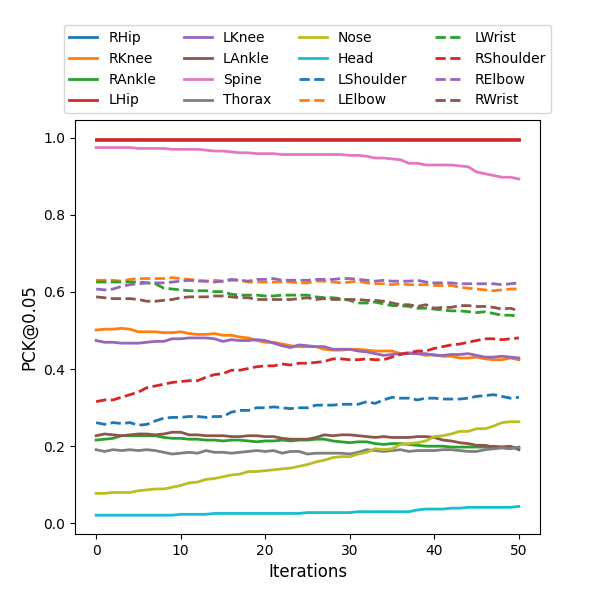} \label{fig:joint_corr_1}} \\
    \subfloat[PCK@0.1]{\includegraphics[width=0.8\linewidth]{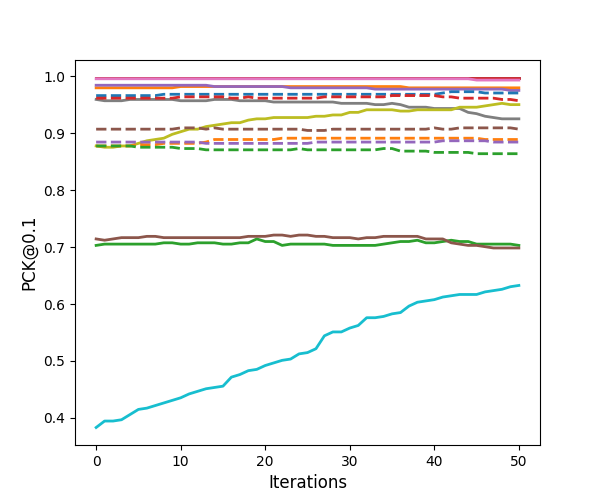} \label{fig:joint_corr_2}} \\
    \subfloat[PCK@0.2]{\includegraphics[width=0.8\linewidth]{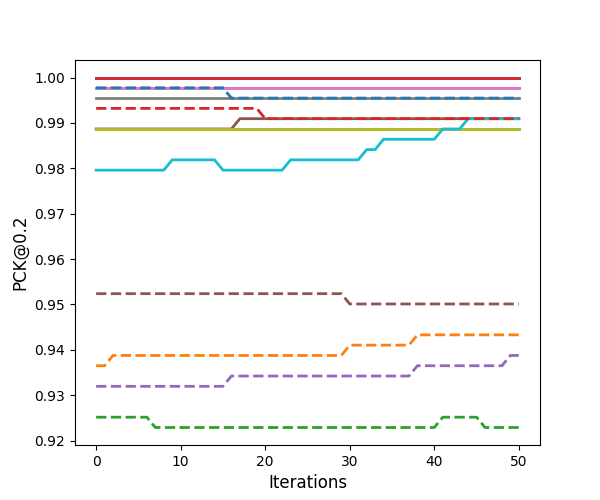} \label{fig:joint_corr_3}} 
    \caption{Mean PCK per joint over 50 iterations.}
    \label{fig:joint_corr}
\end{figure}

To inspect the pose correction results in greater detail, we conducted a joint-wise performance analysis, as shown in Figure \ref{fig:joint_corr}. The mean PCK@0.05, 0.1, and 0.2 values per joint across 50 correction iterations are plotted. Each subplot contains 16 curves, each corresponding to a specific joint.
At the PCK@0.05 threshold, the most effectively corrected joints include RShoulder (right shoulder), LShoulder (left shoulder), and Nose, while joints such as Head, Thorax, RAnkle (right ankle), and LAnkle (left ankle) exhibit substantial errors that are not well corrected.
At the PCK@0.1 level, Head, RAnkle, and LAnkle remain the most error-prone joints, with only the Head showing significant improvement during correction.
At the PCK@0.2 threshold, LWrist (left wrist) and RWrist (right wrist) show large initial errors with minimal improvement, whereas RElbow (right elbow) and LElbow (left elbow) demonstrate notable error reduction over iterations.
These observations suggest that joints exhibiting less variation in motion (i.e., smaller movement ranges) are corrected more effectively by the NDF pose prior. In contrast, joints with greater motion complexity, such as wrists and ankles, are more challenging to correct. This indicates that additional training data, particularly with a wider range of motion patterns, may help improve the model's ability to refine such joints more accurately.

\subsection{Ablation studies}
\label{subsec:exp_ablation}
To examine the individual contributions of our proposed components, we conducted a series of ablation studies. Specifically, we focused on four key aspects of our method: (1) the polar coordinate representation, (2) the batch projection-based data augmentation, (3) the arc-radius distance metric, and (4) the gradient loss term defined in Equation \ref{eq:loss_grad}. To this end, we performed experiments across five different configurations to isolate the impact of each component:
\begin{itemize}
    \item \textbf{Polar baseline}: This baseline setting uses the polar coordinate-based representation for 2D poses. The distances of fake poses are defined using the arc-radius distance, as described in Equation \ref{eq:polar_dist}. The model is trained using the batch projection-based data augmentation process with both distance-related loss and gradient-related loss.  
    \item \textbf{Angular baseline}: This baseline setting follows the angular representation as described in \cite{raychaudhuri2023prior}. The distances of fake poses are defined using the angular distance. The model is trained using both batch projection-based data augmentation and gradient-related loss.  
    \item \textbf{Polar w/o bp}: This setting is similar to the polar baseline, except that batch projection augmentation is not applied during training. To compensate for this, the model is trained for additional epochs until no further improvement is observed on the validation set.
    \item \textbf{Polar w/o grad. loss}: This setting is similar to the polar baseline, except that the gradient loss term is excluded during training.
    \item \textbf{Polar w/o AR. dist.}: This setting is similar to the polar baseline, except that the distances of the fake poses are defined using the geodesic distance (Equation \ref{eq:geodesic_dist}).
\end{itemize}
\vspace{0.5em}

The ablation study is conducted on pose detections from AlphaPose, using three PCK metrics. The results are summarized in Table \ref{tab:ablation}. The polar baseline achieves the highest accuracy in PCK@0.2, ranks second in PCK@0.1, with only a 0.01\% lower accuracy than the best, and is third in PCK@0.05. Therefore, the polar baseline demonstrates the best overall performance.

By comparing the polar baseline with the angular baseline, we observe that the polar representation offers slightly better performance. The results from 'Polar w/o bpj' indicate that the batch sample projection plays a key role in enhancing model performance. The 'Polar w/o grad. loss' yields the lowest performance across all three metrics, suggesting that this loss term significantly influences the learning of a robust NDF prior. Lastly, the 'Polar w/o AR dist.' achieves the second-best accuracy in PCK@0.05 and PCK@0.2, demonstrating that while the geodesic distance performs slightly worse than the proposed arc-radius distance, the difference is not substantial.

\begin{table}
    \centering
    \setlength{\tabcolsep}{4pt}
    \begin{tabular}{@{}l c c c c@{}}
    \hline
    Setting & PCK@0.05 & PCK@0.1 & PCK@0.2\\
    \hline
    Polar baseline &45.11 &\underline{80.10}  &\textbf{91.62} \\
    %\cline{2-3}
    Angular baseline & 44.96 &\textbf{80.11}  &91.53  \\
    %\cline{2-3}
    Polar w/o bp &\textbf{45.20}  &79.99  &91.45  \\
    %\hline
    Polar w/o grad. loss &44.91 &80.02 &91.40  \\
    %\hline
    Polar w/o AR. dist. &\underline{45.15} &80.06 &\underline{91.58}  \\
    \hline
    \end{tabular}
    \caption{Ablation studies on AlphaPose detections}
    \label{tab:ablation}
\end{table}

\subsection{Further studies}
\label{subsec:further_studies}
We conducted an additional analysis to evaluate the convergence of the pose prior. As shown in Figure \ref{fig:joint_corr}, the accuracy of some joints decreases over iterations, suggesting that the manifold learned by the NDF pose prior did not fully converge. In this analysis, we compare the baseline performance with the average of the highest accuracy achieved for each pose sample using the same NDF pose prior. The results presented in Table \ref{tab:convergence} demonstrate that while the model achieves significantly improved pose corrections during the refinement process, it does not consistently converge to the optimal correction. This indicates that even higher performance could be attained if the convergence issue were effectively addressed.

\begin{table}
    \centering
    \begin{tabular}{@{}l c c c c c@{}}
    \hline
    Model& Method & PCK@0.05 & PCK@0.1 \\
    \hline
    AlphaPose  & baseline & 45.11 & 80.10  \\
    & {opt. converged} &50.31 &82.05 \\
    \hline
    HRNet & baseline &41.39 &76.35  \\
    & {opt. converged} &46.03 &79.33 \\
    \hline
    ViTPose  & baseline &51.78 &90.52  \\
    & {opt. converged} &57.06 &92.34 \\
    \hline
    \end{tabular}
    \caption{Assessment of the convergence behavior of the NDF pose prior. Baseline performances are compared with the average of the highest accuracy achieved for each pose during the correction process, reported as \textit{opt. converged} in the table.}
    \label{tab:convergence}
\end{table}
\section{Discussions and conclusions}
\label{sec:discussions}

In this work, we proposed a polar coordinate-based representation for learning 2D pose prior with the neural distance field. To facilitate the training of the pose prior, we also proposed gradient loss term and an arc-radius distance for polar coordinate representations. To enable model training with small datasets, we proposed a data augmentation method to generate new sample batches by gradient-based projection. The experimental results showed a consistency improvement of the pose key point accuracy, regardless of the source model for the pose estimations. We also show that the model could be trained with less than a quarter of the original training data (280 poses out of 1249 poses) without a big sacrifice of performance. 

Despite the effectiveness and robustness of the proposed method, there are still two major issues. The first is that the model tends to be less capable of correcting the large motion joints (wrists and ankles), which could be caused by insufficiency of data varieties. Moreover, currently, the model still has difficulty with convergence, which greatly hinders the performance. This article shows that the model has potential to achieve a much better result if the convergence issue were properly resolved.
{
    \small
    \bibliographystyle{ieeenat_fullname}
    \bibliography{main}
}


\end{document}